\documentclass[pdflatex,sn-mathphys-num,oneside]{sn-jnl}%

\usepackage{graphicx}
\usepackage{multirow}
\usepackage{amsmath,amssymb,amsfonts}
\usepackage{amsthm}
\usepackage{mathrsfs}
\usepackage[title]{appendix}
\usepackage{xcolor}
\usepackage{textcomp}
\usepackage{manyfoot}
\usepackage{booktabs}
\usepackage{algorithm}
\usepackage{algorithmicx}
\usepackage{algpseudocode}
\usepackage{listings}
\usepackage{colortbl}
\usepackage{tikz}
\usepackage{subcaption}
\usepackage{overpic}
\usetikzlibrary{positioning}
\usepackage[acronym,automake]{glossaries}
\makeglossaries

\usepackage[left]{lineno}
\newacronym{fbp}{FBP}{Filtered Back Projection}
\newacronym{gd}{GD}{Gradient Descent}

\begin{document}

\title[Article Title]{Advanced Brain Tissue Imaging via Data-Consistent Diffusion Priors in Laminographic X-Ray Nanoimaging}


\author[1,2]{\fnm{Wenxuan} \sur{Fang}}\email{wenxuan.fang@epfl.ch}

\author[2]{\fnm{Abraham} \sur{L. Levitan}}
\author[2]{\fnm{Ana} \sur{Diaz}}
\author[3]{\fnm{Carles} \sur{Bosch}}
\author[2]{\fnm{Adrian} \sur{Wanner}}
\author[3]{\fnm{Andreas} \sur{T. Schaefer}}
\author[2]{\fnm{Mirko} \sur{Holler}}
\author[2]{\fnm{Tomas} \sur{Aidukas}}
\author[2,5]{\fnm{Nicholas} \sur{W. Phillips}}
\author[3]{\fnm{Yuxin} \sur{Zhang}}
\author[4]{\fnm{Alexandra} \sur{Pacureanu}}
\author*[1,2]{\fnm{Manuel} \sur{Guizar-Sicairos}} \email{manuel.guizar-sicairos@psi.ch}
\author*[2]{\fnm{Luis} \sur{Barba}}
\email{luis.barba-flores@psi.ch}


\affil*[1]{\orgname{École Polytechnique Fédérale de Lausanne}, \orgaddress{\city{Lausanne}, \country{Switzerland}}}

\affil[2]{\orgname{Paul Scherrer Institute}, \orgaddress{\city{Villigen}, \country{Switzerland}}}

\affil[3]{\orgname{Sensory Circuits and Neurotechnology Lab, The Francis Crick Institute}, \orgaddress{\street{1 Midland Road}, \city{London}, \postcode{NW1 1AT}, \country{UK}}}

\affil[4]{\orgname{ESRF, The European Synchrotron}, \orgaddress{\city{Grenoble}, \country{France}}}

\affil[5]{\orgname{Mineral Resource CSIRO}, 
\orgaddress{\street{Clayton}, \city{Victoria}, \country{Australia}}}

\abstract{
Nanoscale imaging of mammalian brains is critical for connectomics. X-ray laminography enables high-throughput imaging of extended, plate-like biological specimens. However, the tilted acquisition geometry leads to incomplete Fourier-space coverage, giving rise to a missing-cone of information.
Conventional reconstruction methods cannot recover unmeasured information within the cone, resulting in artifacts that distort fine brain structures. While resolving these requires modeling 3D structure, direct 3D deep learning approaches are limited by data scarcity and computational cost.
Here we introduce LUCID (Laminography with Unified Consistent Diffusion), a framework that combines multi-view diffusion priors with projection-domain data consistency. LUCID integrates complementary 3D structural information while enforcing strict alignment with the laminography forward model.
On simulated datasets, LUCID substantially improves spatial fidelity and restores missing Fourier components, outperforming baseline methods.
Applied to experimental laminography data, LUCID generalizes robustly despite being trained exclusively on fully sampled tomographic volumes, and effectively recovers unmeasured Fourier information.}


\maketitle

\section{Introduction}\label{sec1}

The nanoscale architecture of the brain's connectome, a comprehensive map of neural circuitry, holds the key to deciphering the structural underpinnings of cognition, behavior, and neurological disorders~\citep{abbott2020mind,helmstaedter2026synaptic}. To recover the connectome, the synapic connections between neurons must be resolved, which requires resolutions on the order of 10–20 nm. At this resolution, synaptic features become resolvable, enabling
the identification of synaptic interfaces and providing critical information about the wiring principles of the central nervous system. However, achieving three-dimensional (3D) imaging at this resolution across millimeter-scale tissue volumes remains a formidable challenge. While electron microscopy (EM)-based techniques, such as serial sectioning and focused ion beam/scanning electron microscopy (FIB/SEM), provide a resolution sufficient to reliably identify synapses~\citep{knott2008serial,merchan2009counting}, their reliance on thin-sectioning or ablation introduces artifacts and limits scalability.
X-ray imaging provides a non-destructive alternative, and recent advances in synchrotron-based ptychography have demonstrated nanometric resolution in a wide variety of specimens of tens of microns in diameter~\citep{dierolf2010ptychographic, yu2018three, holler2019three, aidukas2024high,beck2025situ}. While nanoscale X-ray imaging can resolve synaptic structures~\citep{bosch2025nondestructive, laugros2025self}, conventional X-ray computed tomography (CT) relies on rotation around an axis perpendicular to the X-ray beam. Applying this geometry to extended brain tissue would require cutting cuboid-like samples to allow full angular coverage. However, partitioning large brain volumes into such geometries without disrupting or losing delicate neural structures is generally not feasible. In contrast, sectioning brain tissue into thick slices is experimentally practical and preserves structural integrity over large fields of view. 

In contrast, sectioning brain tissue into thick slices is experimentally practical and preserves structural integrity over extended fields of view. X-ray laminography~\citep{nikitin2024laminography}, with its tilted imaging geometry, is naturally suited to such plate-like specimens and avoids the need for cylindrical sample preparation. Although the preparation strategy resembles serial sectioning in electron microscopy, X-rays can penetrate substantially thicker tissue sections, enabling higher-throughput volumetric imaging of neural tissue.

Computed laminography (CL) employs a tilted rotation axis forming an angle smaller than $90^\circ$ with respect to the X-ray beam. Compared with limited-angle tomography, laminography provides more uniform effective thickness across projection angles and reduces the extent of missing information in Fourier space. In limited-angle tomography, incomplete angular coverage gives rise to a missing wedge in reciprocal space, leading to severe directional artefacts and anisotropic resolution degradation~\citep{arslan2006reducing,bartesaghi2008classification}. By contrast, the tilted acquisition geometry of laminography distributes the missing information into a biconical region rather than a wedge, resulting in more balanced sampling and improved structural preservation across views. These properties make laminography particularly well suited for high-resolution 3D imaging of thick neural sections.

\begin{figure}[h]
  \centering
  \includegraphics[width=1\textwidth]{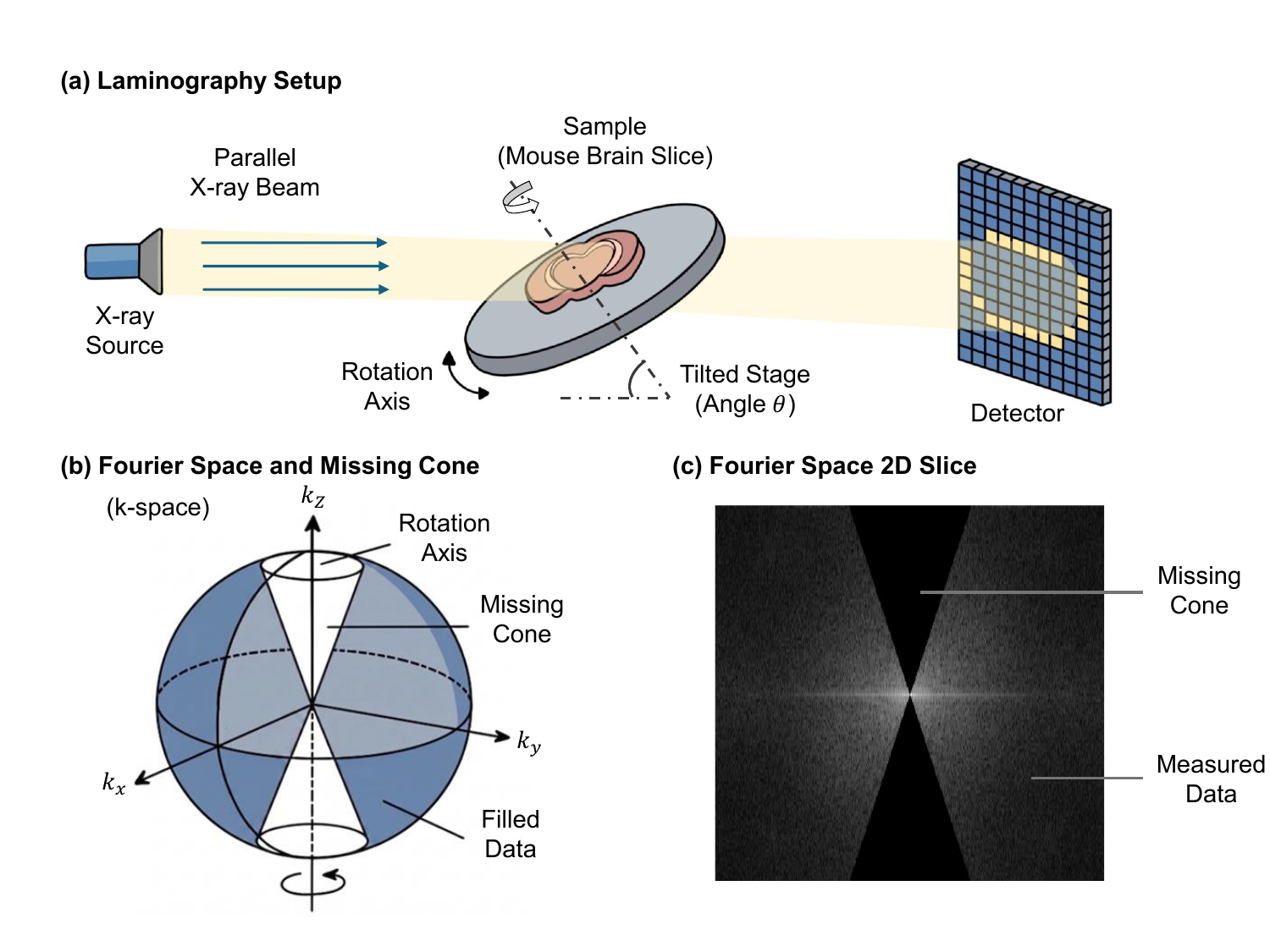}
  \caption{
\textbf{Laminography acquisition geometry and the missing-cone problem.}
\textbf{(a)} Schematic of X-ray laminography with a plane-wave beam. A parallel X-ray beam illuminates a tilted specimen, which rotates around an axis inclined by an angle $\theta$ with respect to the beam. The transmitted intensity is recorded by a pixelated detector. 
\textbf{(b)} Corresponding sampling pattern in 3D Fourier space ($k$-space). The tilted rotation geometry leads to incomplete coverage, leaving a biconical region (missing cone) unsampled along the rotation axis. 
\textbf{(c)} Representative 2D Fourier slice illustrating the missing-cone region (dark wedge), which results in anisotropic frequency loss. This incomplete sampling fundamentally limits reconstruction fidelity and induces artifacts. It should be noted that (a) is provided as a schematic to explain the concept of laminography, but the experimental data shown in this paper was not acquired with a parallel beam geometry but rather using ptychographic X-ray laminography \cite{holler2019three}.
}
  \label{artifacts}
\end{figure}

Despite the aforementioned advantages, the imaging geometry of laminography gives rise to an ill-posed reconstruction problem. As illustrated in Fig.~\ref{artifacts}a, the rotation axis is tilted with respect to the incident X-ray beam, which leads to an incomplete coverage of the 3D frequency domain, leaving a characteristic biconical region unsampled along the rotation axis~\citep{holler2019three,nikitin2024laminography} as shown in Fig.~\ref{artifacts}b. This missing-cone geometry induces a highly anisotropic transfer of information, as further visualized in a 2D Fourier slice in Fig.~\ref{artifacts}c, where entire regions of Fourier space are absent. This geometry results in limited resolution along the axis normal to the sample plane~\citep{du2025x}, manifesting as a pronounced axial elongation, interlayer aliasing, and directional blurring. Fine neural features are therefore smeared across adjacent layers,  becoming so distorted as to be no longer visible. Addressing this loss of information along the vertical axis is essential for achieving faithful three-dimensional reconstruction in laminography. Fig. \ref{artifacts} provides a conceptual schematic of the laminography acquisition geometry. It is intended to illustrate the principle of laminography and does not represent the specific experimental setup, where synaptic-resolution data were acquired using ptychographic X-ray laminography.

CL reconstruction is commonly performed using a modified or extended version of the \gls{fbp} algorithm~\citep{helfen2009phase}. Unfortunately, because \gls{fbp} provides no means to compensate for the missing cone, it is unable to correct for missing-cone related artifacts. Therefore, most existing efforts to correct for such artifacts have focused on iterative reconstruction strategies. These approaches can partially mitigate this issue by roughly estimating the missing information. They do this by minimizing a data error metric using algebraic~\citep{que2012computed} or regularized iterative solvers~\citep{lu2023anisotropic}. To further incorporate prior knowledge, anisotropic and edge-preserving regularization schemes~\citep{kang2023accelerated} have been introduced to suppress inter-slice aliasing and reduce noise. Although effective in controlled settings, such methods require careful tuning of regularization terms and hyperparameters, and often lead to oversmoothing of fine structural details. 

A closely related problem in tomography is the missing-wedge artifacts encountered in limited-angle acquisition geometries, particularly in cryo-electron tomography (cryo-ET). To mitigate these effects, several learning-based reconstruction approaches have been proposed, including GAN-based frameworks~\citep{ding2019joint} operating jointly in sinogram and tomographic domains, as well as more recent iterative and self-supervised methods such as IsoNet~\citep{liu2022isotropic} and DeepDeWedge~\citep{wiedemann2024deep}. These approaches demonstrated the potential of learned priors to recover missing structural information. Similarly, deep learning–based methods have recently been explored for CL reconstruction and mitigation of missing cone artifacts. Some propose post-processing networks applied after reconstruction~\citep{zou2025artifact}. These approaches remain decoupled from the reconstruction and do not enforce data consistency, as the denoised outputs are not re-constrained by the measured projections. Consequently, they primarily act as artifact-correction modules rather than generative models capable of recovering genuinely missing information. The approach in~\citep{liu2026laminodiff} introduced a generative network that relies on an unattainable dual-scale CT-CL fusion strategy to construct artifact-free ground truth (GT) labels for training. Acquiring such perfectly aligned, high-fidelity fusion targets is physically intractable for intact biological tissues.

The missing-cone problem in laminography is intrinsically 3D, and in principle requires processing the full volumetric data to recover lost information. In contrast to conventional tomography, due to the tilted geometry of the rotation axis, the laminography reconstructions and therefore also the recovery of missing-cone information cannot be split into 2D slices.  However, direct 3D learning-based reconstruction faces substantial practical barriers. Volumetric data is inherently scarce, and operating on high-resolution 3D volumes demands prohibitively large computational resources. Both training and inference that operate on complete volumes become slow or simply infeasible, making fully 3D approaches difficult to deploy in realistic imaging workflows. Existing methods for laminography that resort to independent slice-by-slice 2D reconstructions~\citep{zou2025artifact, liu2026laminodiff} are inadequate, because they inevitably sacrifice volumetric consistency and fail to capture structural dependencies across the volume.

The existing methods mentioned above, iterative, regularized, or learning-based, have been developed and validated primarily in industrial settings that involve structures with limited diversity and heterogeneity, such as printed circuit boards or integrated circuits~\citep{que2012computed, lu2023anisotropic, jia2023multi,ghandourah2023evaluation, xu2012comparison, zou2025artifact,liu2026laminodiff}. In contrast, brain tissue exhibits extreme structural heterogeneity, with diverse cell types, complex neuropil textures, and densely interconnected synaptic networks. These characteristics introduce unique reconstruction challenges, making it especially hard to exploit the information which does exist about their prior distribution. 

Diffusion models~\citep{yin2024survey, song2020score, song2019generative, lee2023improving} have shown remarkable capability in image restoration and reconstruction, suggesting strong potential for addressing the missing cone problem. Yet their implementation in laminography reconstruction, as well as their application to a challenging field like 3D brain imaging remain largely unexplored.

Here, we propose LUCID (Laminography with Unified Consistent Diffusion), a physics-guided generative framework that unifies the strengths of data-driven diffusion priors and measurement-consistent optimization. Rather than operating independently on individual slices or relying on post processing, LUCID employs a multi-axial diffusion strategy across axial, sagittal, and coronal planes, enabling complementary structural learning. Crucially, the diffusion model is embedded directly within the iterative laminography reconstruction loop, allowing the generative prior to evolve jointly with projection-domain data-consistency updates. This tight coupling reconciles learned anatomical priors with the governing physics of the acquisition process, steering reconstructions toward physically faithful and anatomically plausible solutions.

In this work, we first introduce the LUCID framework and evaluate its performance on simulated laminography datasets generated from high-fidelity tomographic volumes, which enable quantitative analysis of artifacts and restoration of the missing cone. In parallel, we establish an experimental benchmark by acquiring, curating, and publicly releasing what is, to our knowledge, the first nanoscale ptychographic X-ray laminography dataset of mouse brain tissue. We then demonstrate robust cross-domain generalization on real experimental laminography data. Across both simulated and experimental settings, LUCID consistently restores structural information, reduces interlayer aliasing, and recovers fine features that are severely degraded by conventional methods. We observe better performance on simulated data, presumably due to substantial differences in acquisition and resolution between the training data and the experiment data. Finally, we assess the reliability and adherence to the GT. The recovered structures remain tightly constrained by the measured projections, indicating that the generative expressivity of the diffusion prior is effectively regulated by physical data consistency.


\section{Results}\label{sec2}

\subsection{LUCID framework}
\begin{figure}[h]
  \centering
  \includegraphics[width=1\textwidth]{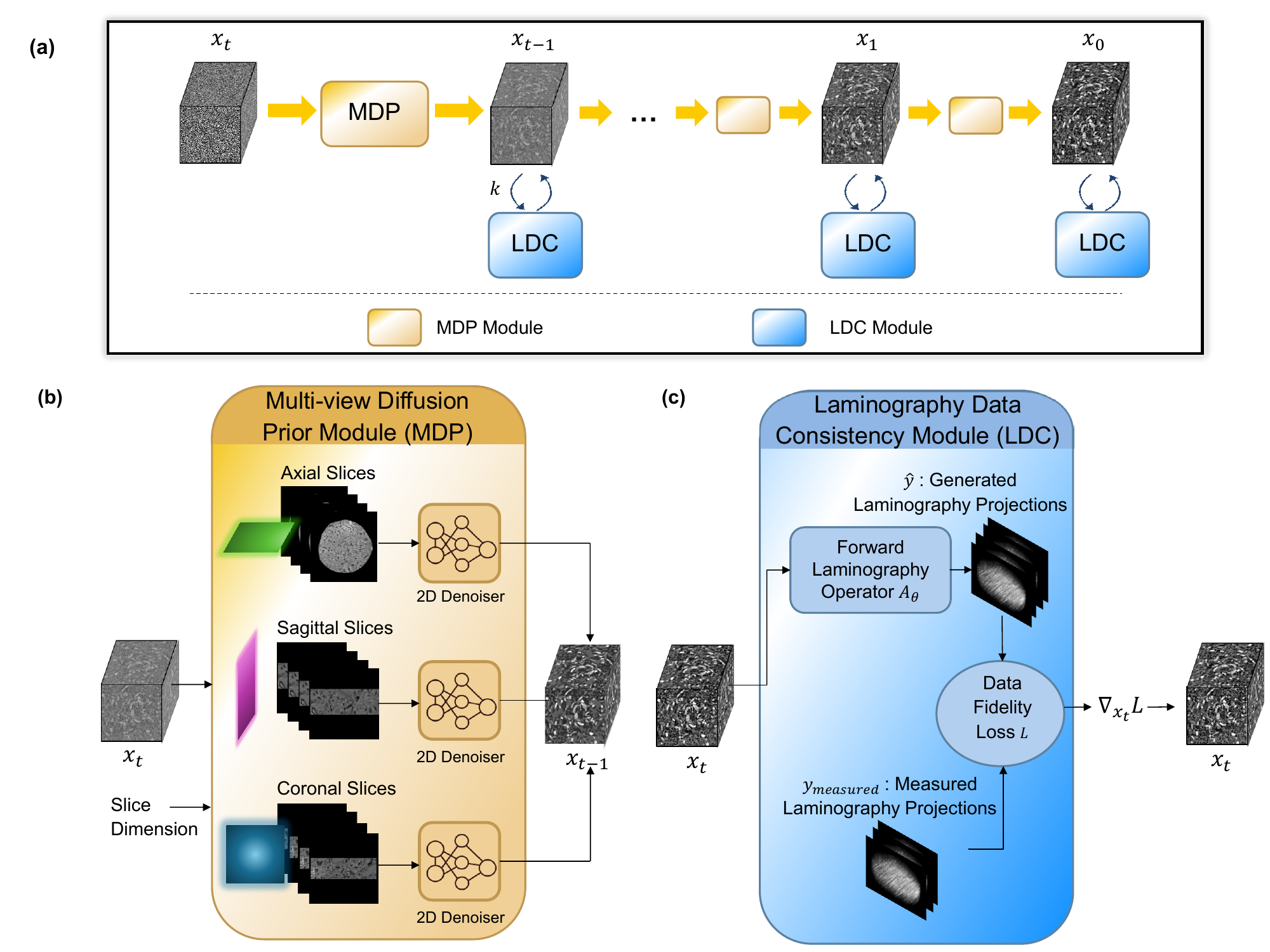}
  \caption{
\textbf{Overview of the LUCID reconstruction framework.}
The reconstruction is formulated as an iterative process that alternates between a learned generative prior and laminography data consistency. 
\textbf{(a)} LUCID pipeline. Starting from Gaussian noise $x_T$, the volume is progressively refined through multiple denoising steps (MDP), interleaved with laminography data consistency updates, yielding the final reconstruction $x_0$. 
\textbf{(b)} Multi-view Diffusion Prior Module (MDP). The current 3D estimate $x_t$ is decomposed into axial or sagittal, or coronal slices, each processed by a dedicated 2D diffusion denoiser to restore anatomically plausible structures. 
\textbf{(c)} Laminography Data Consistency Module (LDC). The volume is projected using the forward operator $\mathcal{A}_\theta$ to generate synthetic projections $\hat{y}$, which are compared with measured projections $y_{\mathrm{measured}}$ via a data fidelity loss to enforce consistency with physical measurements. 
The two modules are coupled in an iterative loop, ensuring that reconstruction remains both measurement-consistent and anatomically coherent.
}
  \label{pipeline}
\end{figure}

We introduce LUCID, a physics-guided generative reconstruction framework that tightly integrates multi-view diffusion priors with laminography data consistency. As illustrated in Fig.~\ref{pipeline}, LUCID is formulated as an iterative refinement process, in which a 3D volume is progressively updated by alternating between two modules.

\textbf{Training of the diffusion prior.}
To learn a high-fidelity anatomical prior, we train diffusion models on tomography reconstructions~\citep{bosch2025nondestructive}, which provide isotropic, high-resolution 3D volumes under a well-conditioned imaging geometry. Each volume is decomposed into axial, sagittal, and coronal slices, enabling efficient learning of volumetric structures using 2D models. 

For each view, a denoising diffusion model is trained to learn the reverse of a fixed forward noising process, which gradually perturbs clean samples $x_0$ into Gaussian noise $x_T$. The model is optimized to predict and remove noise at each step, thereby capturing the distribution of realistic structures across orientations.

\textbf{Iterative reconstruction with coupled modules.}
At inference, reconstruction is initialized from Gaussian noise $x_T$ and proceeds through a sequence of alternating updates, as depicted in Fig.~\ref{pipeline}a. At each iteration $t$, the current estimate $x_t$ is refined by two complementary operations:

\emph{(i) Multi-view Diffusion Prior Module (MDP).}
The current 3D volume $x_t$ is decomposed into axial, sagittal, and coronal slices in Fig.~\ref{pipeline}b. Each view is processed by its corresponding 2D diffusion denoiser, which removes noise and restores anatomically plausible structures. The denoised slices are then reassembled into an updated 3D estimate $x_{t-1}$. This step primarily acts as a learned prior, promoting structural coherence and compensating for missing information.

\emph{(ii) Laminography Data Consistency Module (LDC).}
The updated volume is passed to the LDC module in Fig.~\ref{pipeline}c, where data consistency is enforced. Specifically, the current estimate is projected using the laminography forward operator $\mathcal{A}_\theta$ to generate simulated projections $\hat{y}$. These are compared with the measured projections $y_{\mathrm{measured}}$ via a data fidelity loss $\|\hat{y} - y_{\mathrm{measured}}\|_2^2$. The resulting gradient is used to update the volume via a gradient descent step with a fixed step size. In practice, this update is performed for a small number of inner iterations within each diffusion step, ensuring consistency with the measurements.

\textbf{Coupled refinement.}
The MDP and LDC modules are applied sequentially and iteratively throughout the reverse diffusion process in Fig.~\ref{pipeline}a, forming a closed-loop optimization. The diffusion prior provides strong regularization by restricting solutions to the manifold of realistic brain structures, while the data consistency module anchors the reconstruction to physically observed data. 

\subsection{Performance on a simulated laminography dataset}

\begin{figure}[h]
  \centering
  \includegraphics[width=1\textwidth]{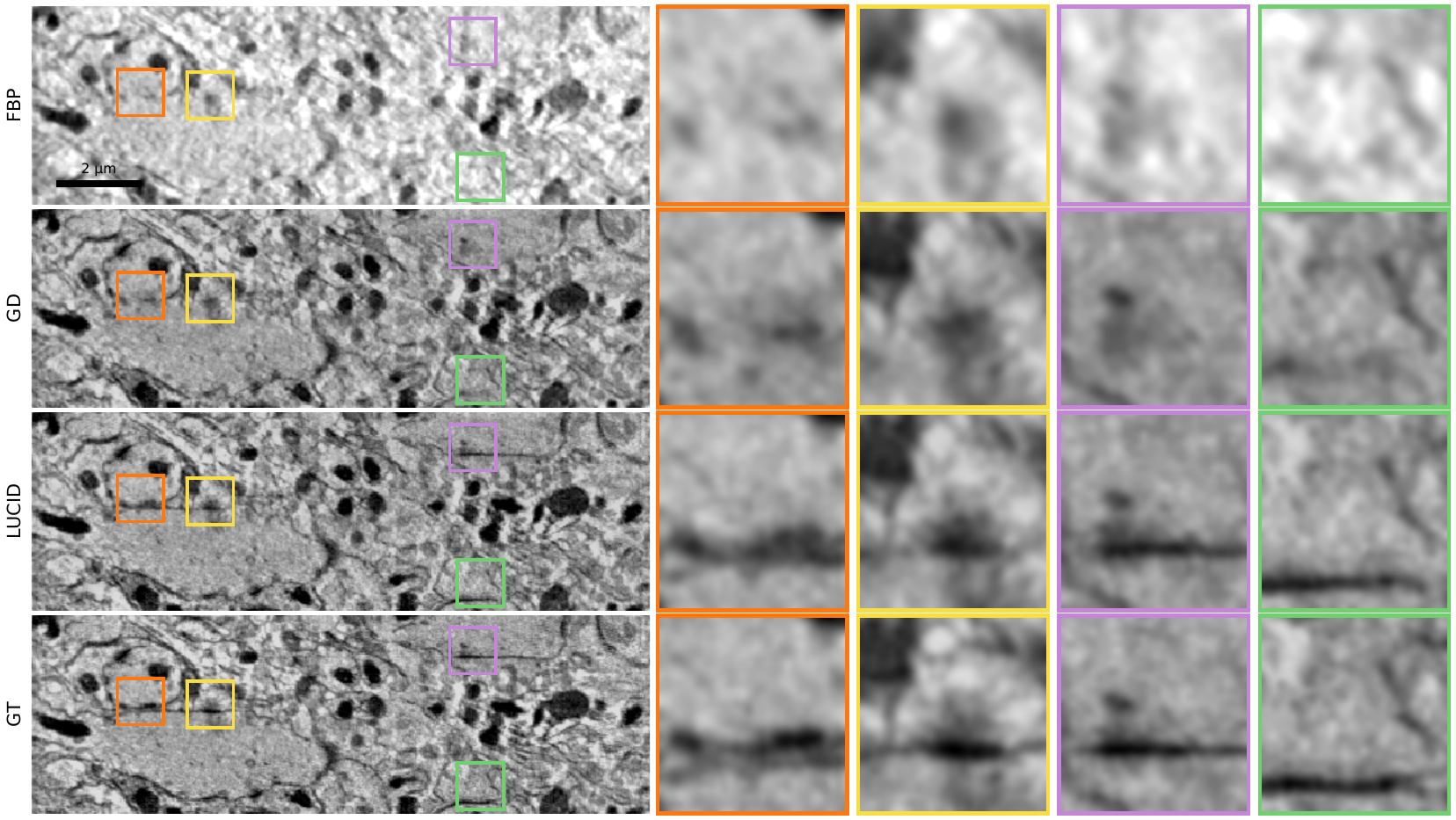}
  \caption{\textbf{Qualitative comparison of reconstruction fidelity on representative brain regions.}
Left: reconstructed slices from \gls{fbp}, GD, LUCID, and GT. Colored boxes indicate regions of interest (ROIs), with corresponding zoom-ins shown on the right. 
\gls{fbp} suffers from severe blurring and elongation due to missing-cone artifacts, while \gls{gd} partially improves structural sharpness but retains anisotropic distortions. In contrast, LUCID recovers fine-scale features with improved contrast and continuity, closely matching the GT. 
Highlighted ROIs correspond to biologically relevant structures: \textbf{orange} and \textbf{yellow} boxes indicate putative synapses, which are poorly resolved in \gls{fbp} and \gls{gd} but clearly recovered in LUCID; \textbf{purple} and \textbf{green} boxes indicate putative plasma membranes, whose continuity and boundary definition are substantially improved by LUCID.}
  \label{sim}
\end{figure}

\begin{figure}[h]
  \centering
  \includegraphics[width=1\textwidth]{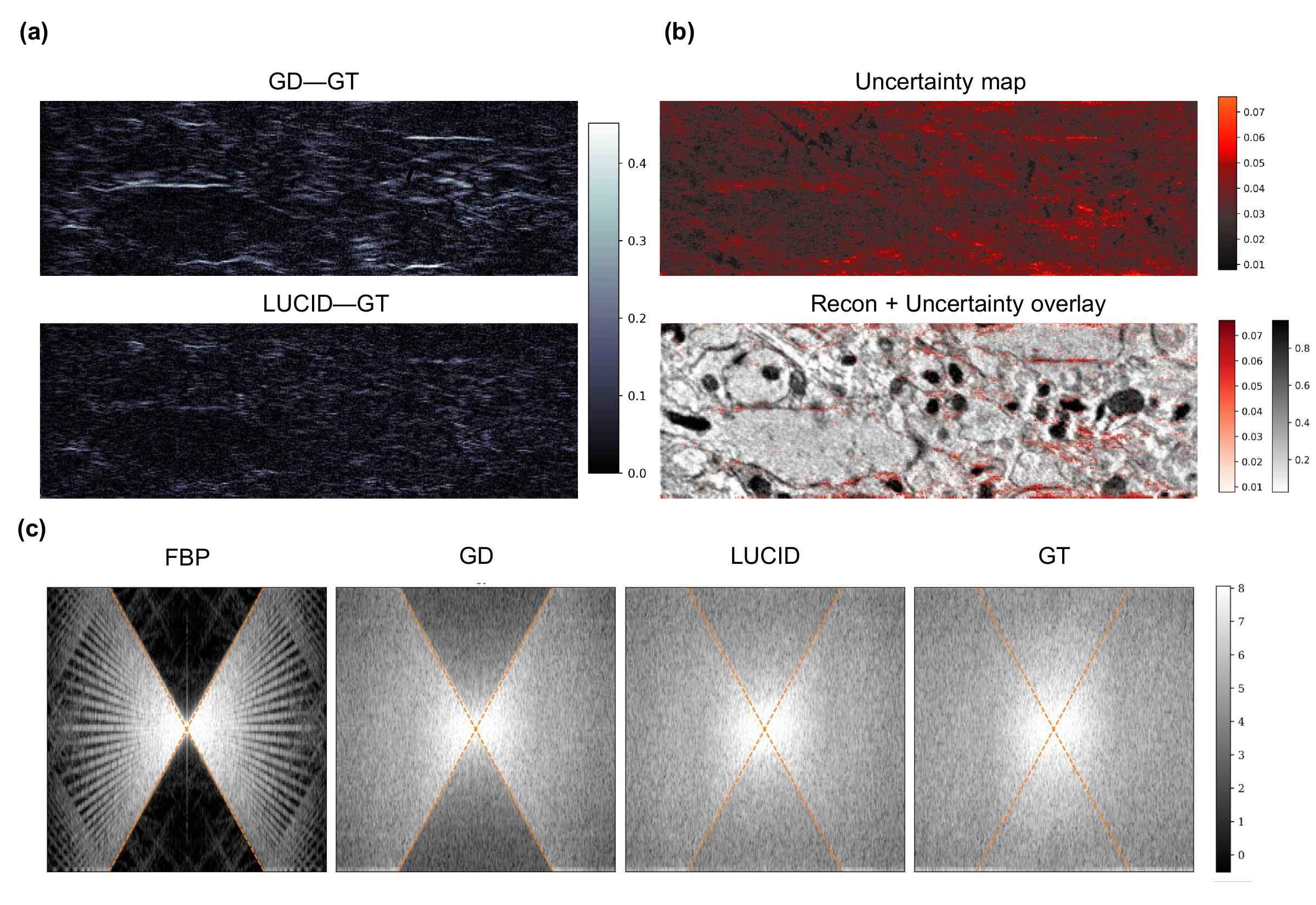}
  \caption{
\textbf{Quantitative and spectral evaluation of reconstruction quality and uncertainty.}
\textbf{(a)} Spatial-domain error maps, i.e. difference to GT, for \gls{gd}, and LUCID. Conventional methods exhibit strong anisotropic errors aligned with the missing-cone direction, whereas LUCID substantially reduces both the magnitude and spatial extent of residuals.
\textbf{(b)} Uncertainty estimation of LUCID reconstructions. \textit{Top:} Voxel-wise uncertainty map computed from multiple stochastic reconstructions, highlighting elevated uncertainty in regions affected by missing-cone information loss. \textit{Bottom:} Overlay of the reconstruction and uncertainty map, showing that regions of high uncertainty spatially coincide with structurally ambiguous areas.
\textbf{(c)} Fourier-domain analysis. Slices of the 3D Fourier magnitude are shown for \gls{fbp}, \gls{gd}, LUCID, and GT, with the missing-cone region delineated by dashed lines. \gls{fbp} and \gls{gd} exhibit pronounced spectral missing energy in the cone, while LUCID recovers substantially more energy within the missing cone, approaching the GT distribution. All reconstructions are angular undersampling, especially visible in \gls{fbp} outside the missing-cone region.}
  \label{sim_compare}
\end{figure}

We evaluate our method on a simulated laminography dataset generated using the laminography forward model applied to high-resolution full-angle tomography volumes, which serve as the ground truth for quantitative evaluation.
The volume used for simulation is strictly held out from training, ensuring a separation between training and evaluation data and enabling an unbiased assessment of generalization performance.

\textbf{Qualitative Evaluation}:
Fig.~\ref{sim} presents a qualitative comparison of reconstruction results obtained with \gls{fbp}, gradient-descent reconstruction (\gls{gd}), and the proposed LUCID framework against the GT. The differences are evident both at the global structural level and in localized regions of interest (ROIs).

\gls{fbp} exhibits pronounced missing-cone artifacts, leading to strong axial blurring and loss of structural contrast. Fine cellular features are largely obscured, and synaptic-like structures in the axial direction are indistinguishable. \gls{gd} partially mitigates these degradations through iterative refinement, yielding improved sharpness; however, residual anisotropic artifacts and inconsistent local contrast persist, particularly along the missing-cone direction.
In contrast, LUCID substantially enhances structural fidelity and continuity. As highlighted in the zoomed ROIs, synaptic-like structures (orange and yellow boxes) that are indistinct in \gls{fbp} and \gls{gd} become clearly delineated, with improved contrast and spatial definition. Similarly, putative plasma membranes (purple and green boxes) exhibit enhanced continuity and sharper boundaries, approaching the appearance observed in the GT. 

These results demonstrate that LUCID not only suppresses missing-cone–induced anisotropy but also restores biologically meaningful ultrastructural features, yielding reconstructions that are both visually consistent and anatomically plausible.

Figure \ref{sim_compare}a shows voxel-wise difference maps between the reconstructions and the GT, revealing where reconstruction errors are spatially concentrated. 
When comparing the difference \gls{gd} to the GT, these errors form continuous bright bands, indicating that these discrepancies coincide with membrane structures that typically define synaptic boundaries, implying that their loss critically affects the reliable identification of synapses. In contrast, in the difference map calculated with the LUCID reconstruction, the membrane-aligned residuals largely vanish and the remaining differences are low and approximately isotropic.

\textbf{Quantitative Evaluation}: Next, we quantitatively evaluated our framework's performance on this simulated ptychographic X-ray laminography (PyXL)  dataset using  peak signal-to-noise ratio (PSNR), structural similarity index (SSIM),
and root mean squared error (RMSE) as metrics to assess reconstruction fidelity. Importantly, all evaluations are performed on volumes not seen during training, ensuring a fair evaluation of generalization performance.

Table~\ref{tab:sota} summarizes the quantitative performance of \gls{fbp}, \gls{gd}, and the proposed LUCID framework. Across all spatial-domain metrics, LUCID achieves the highest reconstruction fidelity. Compared with conventional \gls{fbp}, LUCID improves PSNR by more than 13 dB and reduces RMSE by nearly 75\%, reflecting its strong capability to suppress noise and geometric artifacts. Relative to iterative \gls{gd}, LUCID provides consistent gains in both pixel-wise accuracy and perceptual similarity, confirming that the integration of diffusion priors and data-consistency constraints yields more stable and anatomically faithful reconstructions.

\textbf{Uncertainty Evaluation}: To assess the reliability and stability of our reconstruction framework, we further quantify uncertainty in Fig.~\ref{sim_compare}b by analyzing the variability across multiple stochastic runs of the multi-view diffusion model. Specifically, we run the LUCID reconstruction 20 times, and compute a voxel-wise standard deviation volume as an empirical uncertainty map. 

The spatial distribution of uncertainty reveals strong structural consistency with known limitations of laminography. Regions exhibiting the highest uncertainty are not randomly scattered; instead, they form coherent patterns that align with areas affected by missing-cone artifacts. These regions correspond to Fourier-domain directions that are under-constrained by the acquisition geometry. Their elevated uncertainty therefore reflects the model’s awareness of where the projections cannot uniquely determine the solution, forcing the diffusion prior to play a dominant role. In contrast, well-sampled regions—such as high-contrast boundaries, membranes, and features lying near the central plane—show significantly lower uncertainty, indicating high model confidence where the data provides sufficient coverage.

The uncertainty map is further visualized by overlaying the standard deviation onto the reconstructed slice as shown in Fig.~\ref{sim_compare}b bottom. This representation highlights that coherent ultrastructural patterns such as large neurites and dense organelles exhibit low uncertainty, while diffuse textures and elongated structures aligned with the missing cone display markedly higher uncertainty. This contrast confirms that the diffusion model does not simply “hallucinate” structures in ill-posed regions; it also expresses reduced confidence when the solution cannot be reliably inferred from the data.

Importantly, the uncertainty encodes meaningful structure tied to the physical limitations of the imaging geometry. As such, the uncertainty map can serve as a diagnostic tool for downstream neuroanatomical analysis. In tasks such as connectome reconstruction
where microstructural interpretation requires strict reliability, these uncertainty estimates help distinguish between confidently recovered neural processes and regions where additional measurements or complementary priors may be necessary. Moreover, this information can be propagated to downstream segmentation algorithms, providing spatial and directional cues on where feature representations are less certain and should be treated with caution.

\textbf{Fourier Space Comparison:} 
We compare the reconstructed Fourier frequency spectra of \gls{fbp}, \gls{gd}, and LUCID against the GT in Fig.~\ref{sim_compare}c. The \gls{fbp} result exhibits pronounced loss of information within the missing-cone region. Mild angular undersampling artefacts are visible outside the cone due to the finite number of projection angles used in the simulated acquisition (see Dataset composition and usage in the Methods section). \gls{gd} partially recovers these frequencies, but remains anisotropic and incomplete. In contrast, LUCID substantially enhances spectral density within the missing cone, approaching the isotropic distribution of the GT. This enrichment of Fourier-space energy confirms that LUCID effectively fills the missing cone region, restoring spatial frequency content.

To specifically evaluate the effectiveness of our method in addressing the missing-cone problem, 
we computed three Fourier-domain metrics: 
Cone Spectral Fidelity $(\mathrm{CSF}_{\text{cone}})$, 
Cone Energy Difference $(\Delta E_{\text{cone}})$, 
and Cone Energy $(E_{\text{cone}})$, which are described in Section 4.1. 
$\mathrm{CSF}_{\text{cone}}$ quantifies the correlation between the reconstructed and GT spectra within the missing-cone region, 
while $\Delta E_{\text{cone}}$ measures the relative energy discrepancy in this region. Table~\ref{tab:sota} presents complementary frequency-domain analyses. 
LUCID achieves the highest cone spectral fidelity and the largest recovered spectral energy within the missing-cone region, 
highlighting its ability to restore previously unsampled Fourier components. 
The reduced spectral deviation further indicates improved isotropy and a more balanced frequency distribution. 
Taken together, these metrics demonstrate that LUCID not only enhances spatial-domain fidelity but also provides superior Fourier-space completion, 
accurately recovering high-frequency information that conventional methods fail to reconstruct.

\begin{table*}[htbp]
\centering
\renewcommand{\arraystretch}{1.15}
\caption{Quantitative comparison of different reconstruction methods. Best results are shown in bold.}
\label{tab:sota}
\begin{tabular}{lcccccc}
\toprule
Method 
& PSNR $\uparrow$ 
& SSIM $\uparrow$ 
& RMSE $\downarrow$
& $\mathrm{CSF}_{\text{cone}}\uparrow$
& $\Delta E_{\text{cone}}\uparrow$
& $E_{\text{cone}}\uparrow$ \\
\midrule
\gls{fbp} 
& 16.74 
& 0.4783 
& 0.1456 
& 0.3070 
& -0.9633 
& 0.009 \\

\gls{gd} 
& 26.78 
& 0.9787 
& 0.0457 
& 0.6923 
& -0.8877 
& 0.025 \\

LUCID 
& \textbf{30.14} 
& \textbf{0.9902} 
& \textbf{0.0364} 
& \textbf{0.8742} 
& \textbf{-0.5056} 
& \textbf{0.099} \\
\bottomrule
\end{tabular}
\end{table*}

\subsection{Performance on a experimental laminography dataset}
\begin{figure}[h]
  \centering
  \includegraphics[width=0.7\textwidth]{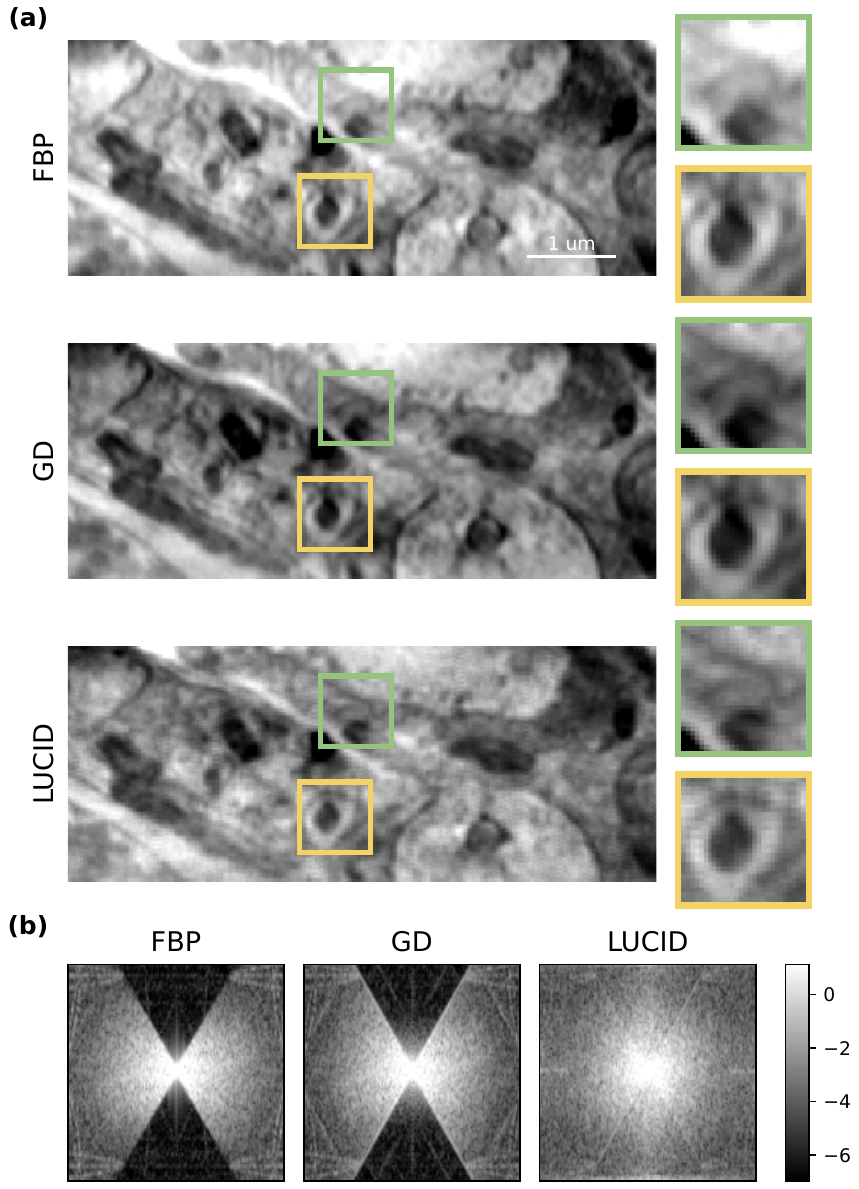}
  \caption{
    \textbf{Reconstruction results on experimental laminography data.}
    \textbf{(a)} Spatial-domain comparison of \gls{fbp}, \gls{gd}, and LUCID reconstructions on real data. Colored boxes indicate ROIs with corresponding zoom-ins shown on the right. \textbf{(b)} Fourier-domain analysis of the corresponding reconstructions. Slices of the 3D Fourier magnitude are shown, with dashed lines indicating the missing-cone region. \gls{fbp} and \gls{gd} display significant spectral gaps, whereas LUCID recovers substantially more energy within the missing cone, resulting in a more isotropic frequency distribution. The reported cone energy quantifies the recovered spectral content within this region.
    }
    \label{real}
\end{figure}

To further validate the capability of LUCID, we evaluate its performance on a real experimental laminography dataset acquired using PyXL~\citep{holler2020lamni}. Unlike the simulated data, the experimental laminography measurements originate from a different domain, and with a distinct spatial resolution. These discrepancies make real-data evaluation a stringent test of model generalization.

Despite being trained exclusively on simulated tomographic data, LUCID generalizes effectively to experimental laminography without any retraining or fine-tuning. Fig.~\ref{real} demonstrates that LUCID restores structural information that is degraded by the missing-cone acquisition geometry in real data.

While \gls{fbp} and \gls{gd} reconstructions suffer from elongation and directional blurring, particularly along linear neural features, LUCID recovers these line-like structures with continuity and contrast. The result in Fig.~\ref{real}a exhibits reduced anisotropy, smoother intensity transitions, and clearer tissue boundaries, showing that LUCID compensates missing information.

This improvement is further supported by Fourier-domain analysis, shown in Fig.~\ref{real}b, where LUCID recovers significantly more spectral energy within the missing-cone region compared to \gls{fbp} and \gls{gd}. The resulting frequency distribution is more isotropic, confirming that the restoration of spatial structures is directly linked to improved recovery of missing Fourier components.


\section{Discussion}\label{sec3}

Here, we present LUCID, a physics-guided generative framework trained on high-quality nanoscale ptychographic-tomography brain volumes and applied to both simulated and real ptychographic laminography data. Our results demonstrate that LUCID effectively restores the information lost due to the missing-cone in laminography, yielding high-fidelity reconstructions with enhanced isotropy and structural continuity. By successfully filling the missing Fourier regions, LUCID overcomes an important limitation in X-ray laminography, enabling faithful recovery of fine neural features such as elongated axonal and dendritic structures which are challenging to reconstruct accurately.

Beyond the proposed reconstruction framework, the first released nanoscale brain laminography dataset represents an important resource for the community. Public datasets have played a central role in accelerating progress in fields such as tomography, cryo-electron microscopy, and connectomics. We anticipate that the availability of experimental laminography data will similarly facilitate the development of reconstruction methods and help establish common benchmarks.

In addition, LUCID produces high-quality volumetric reconstructions that can facilitate downstream brain-imaging studies. In particular, in domains where access to sufficient or diverse training data is restricted, such as due to privacy or acquisition constraints, LUCID’s generative capability provides a data-efficient route to augment existing datasets. This is especially meaningful for neuroscience, for which acquiring large annotated 3D datasets remains technically and ethically challenging. Our findings indicate that integrating diffusion priors with physical constraints significantly enhances the model’s ability to recover coherent neural microarchitecture, providing a robust tool for advancing connectomic analysis and tissue-level interpretation.

To take advantage of these capabilities, we needed to overcome several additional challenges. High-quality 3D generative modeling still demands substantial computational resources, particularly when directly processing volumetric data. To address this, LUCID employs a multi-view diffusion strategy, integrating axial, sagittal, and coronal perspectives. This approach achieves volumetric consistency through efficient 2D inference, greatly reducing computational cost while maintaining 3D coherence, offering a practical balance between accuracy and scalability. Moreover, an inherent domain gap exists between tomography and laminography, as the two modalities differ not only in sampling geometry but also in the statistical distribution of reconstructed voxel intensities, including contrast, dynamic range, and frequency-dependent energy content. To address this challenge, we applied normalization-based domain adaptation, ensuring that LUCID’s outputs remain closely aligned with the physical characteristics of real laminographic data.

Overall, LUCID demonstrates that physics-guided diffusion modeling can bridge modality and resolution gaps in X-ray nanoimaging. The framework provides a data-consistent approach for reconstructing neural tissue imaged with minimally invasive X-ray laminography. By unifying physical modeling and generative intelligence, LUCID establishes a foundation for reliable large-volume brain reconstruction, opening new opportunities for multiscale neuroimaging and other domains facing similar data incompleteness challenges.

\section{Methods}\label{sec3}

\subsection{Dataset and implementation details}

We validate our approach on high-resolution ptychographic X-ray computed tomography (PXCT) datasets of mouse brain tissue acquired at the coherent small-angle X-ray scattering (cSAXS) beamline of the Swiss Light Source (SLS), Paul Scherrer Institute (PSI), Villigen, Switzerland~\citep{bosch2025nondestructive}. The experimental preparation and imaging protocol are designed for structural preservation and contrast at synaptic resolution~\citep{bosch2025nondestructive}, ensuring the dataset serves as a robust benchmark for evaluating reconstruction algorithms.

\textbf{Animals.} Animals used in this study were around 16.4 week old wild-type male mice of C57Bl/6 background . All animal protocols were approved by the Ethics Committee of the board of the Francis Crick Institute and the United Kingdom Home Office under the Animals (Scientific Procedures) Act 1986. All animal IDs are listed in Supplementary Information 1.

\textbf{Sample preparation.} Brain tissue samples with preserved ultrastructure were prepared as described previously~\citep{bosch2025nondestructive}. In brief, mice were sacrificed and a 600 µm-thick horizontal section of the dorsal olfactory bulb extending ~3*3 mm2 was sliced in ice-cold dissecting buffer (phosphate buffer 65 mM, 0.6 mM CaCl2, 150 mM sucrose, 300 ± 20 mOsm/L) using a Leica VT1200S vibratome and quickly transferred to ice-cold fixative (1.25\% glutaraldehyde and 2.5\% paraformaldehyde in 150mM sodium cacodylate buffer, pH 7.40, 300 ± 20 mOsm/L containing 0.02\% sodium azide). After overnight fixation at 4°C the fixative was washed with ice-cold wash buffer (150mM sodium cacodylate buffer, pH 7.40, 300 ± 20 mOsm/L) before being stained with a ROTO protocol~\citep{pallotto2015extracellular} using a Leica automated tissue processor (EMTP). The staining process involved buffered osmium (2\% osmium tetroxide in 150mM sodium cacodylate buffer pH 7.40 for 1h30min at 20°C) followed directly by buffered potassium ferrocyanide (3\% in the same buffer described for the first osmium, for 1h30min at 20°C), 1\% thiocarbohydrazide “TCH” (aq, 45 min at 30°C), 2\% osmium tetroxide (aq, 3h at 20°C), 2\% filtered uranyl acetate (aq, overnight at 4°C, followed by 2h, 50°C) and lead aspartate “LA” (pH 5.5, prepared as in~\citep{walton1979lead}, for 2h 50°C). Six 10 min washes in double distilled water preceded every staining step from TCH onwards, always at 20°C except warmer washes at 50°C  before and after TCH and before LA. Samples were then dehydrated at 20°C by successive washes in ethanol solutions with increased concentration (70\%, 90\%, 90\%, 2x100\%) and transferred to acetonitrile (2x100\%, 30 and 60 min). The TGPAP–DDM resin consists of the tri-functional epoxy resin TGPAP and the hardener DDM at a weight ratio of DDM:TGPAP = 1:2~\citep{bosch2025nondestructive}. DDM was first dissolved in acetonitrile heated to 70 °C and subsequently, TGPAP was added. The samples were incubated in 1:3 resin: acetonitrile for 2 h at room temperature, 1:1 resin: acetonitrile for 2–24 h at room temperature and subsequently the samples were placed in 1:1 resin: acetonitrile and cured for 12–72 h at 80 °C. As the boiling point of acetonitrile is at 82 °C, it is important to keep the sample container lid sufficiently open such that the acetonitrile can evaporate during the curing process.

\textbf{Sample trimming into pillars for PXCT.} Resin-embedded brain samples were trimmed using a diamond knife to the approximate region of interest, as defined by preliminary microCT imaging. Cylindrical pillars of targeted histological layers were extracted using a 30 keV Ga-ion beam of 13 nA on a Zeiss NVision 40 Gallium FIB-SEM at PSI. Samples were mounted onto dedicated PXCT holders using an integrated micromanipulator, with Ga-assisted carbon deposition to ensure mechanical stability. Subsequent fine polishing was carried out in multiple stages, progressively reducing Ga-beam currents from 65 nA to 2.5 nA to achieve smooth, damage-minimized surfaces suitable for coherent X-ray illumination. The preparation of the brain samples is described in detail in~\citep{bosch2025nondestructive}.

\textbf{PXCT instrumentation and acquisition.} 
All PXCT measurements were performed on the OMNY instrument~\citep{holler2018omny} at cSAXS. 
Coherent X-rays of 6.2~keV photon energy, corresponding to a wavelength of 2~\text{\AA} were produced using a fixed-exit double-crystal Si(111) monochromator. 
For most datasets, the illumination optics consisted of a Fresnel zone plate (FZP) of 220~\(\mu\)m diameter and 60~nm outermost zone width with a focal length of 66.0~mm at 6.2 keV, fabricated at PSI, while the last beamtime e19533 employed an FZP from XRNanotech with 250~\(\mu\)m diameter and 30~nm outermost zone width with focal length of 37.5~mm at 6.2 keV. 
In both cases, a gold central stop with 40~\(\mu\)m and an order-sorting aperture with 30~\(\mu\)m  diameter were used to block the unscattered beam, and structured illumination was optimized through locally displaced zones in the FZP. The FZPs were fabricated with designed wavefront aberrations to improve imaging quality~\citep{odstrvcil2019towards}.
A secondary source was defined by a 20~\(\mu\)m horizontal slit placed approximately 12~m from the undulator source, and the sample was positioned slightly downstream of the focal spot, with beam diameters of 8~\(\mu\)m for the first four beamtimes, and 5~\(\mu\)m for beamtime e19533. 
The far-field X-ray diffraction patterns were detected using an invaccum Eiger 1.5M~\citep{guizar2014high} detector placed approximately 7.2~m downstream~\citep{bosch2025nondestructive}.

\textbf{PyXL instrumenation and acquisition.} 
Samples of about 5 micron thickness were cut from metal-stained, dehydrated and resin-embedded brain tissue specimens using a diamond knife and deposited on a silicon nitride membrane for measurement. PyXL measurements on mouse brain were taken with the laminographic nano-imaging (LamNI) instrument~\citep{holler2020lamni} at the cSAXS beamline, of the SLS, PSI, Switzerland. Coherent X-ray photons with energy of 6.2 keV where focused by a FZP of 170 micron diameter and 60 outermost zonewidth fabricated with designed wavefront aberrations \citep{odstrvcil2019towards}. Ptychography scans were taken with a circular FOV of 27 micron diameter on the plane of the sample. On the plane perpendicular to the X-ray propagation the scan followed a Fermat spiral pattern~\citep{huang2014optimization} with an average step size of 0.5 microns. Each ptychogram had 1082 scanning points, at each scanning point an Eiger 1.5M \citep{guizar2014high} measured a far-field diffraction pattern with an exposure time of 0.1 seconds. In total 752 projections were measured at sample orientation angles uniformly distributed between 0 and 360 degrees, yielding a reconstruction pixel size of 27 nm.

\textbf{Ptychographic reconstruction and tomography.} 2D Projection reconstructions from the first four beamtimes were obtained using several hundred iterations of the difference map algorithm followed by maximum likelihood refinement, whereas data from beamtime e19533 were reconstructed with two probe modes and 600 iterations of ML using the ptychoshelves package~\citep{wakonig2020ptychoshelves}. Projections were aligned using a tomographic consistency-based algorithm, followed by modified \gls{fbp} with Hanning or ramp filters~\citep{odstrvcil2019alignment}. Non-rigid corrections were applied to mitigate sample drift and deformation~\citep{odstrcil2019ab}. For the PyXL ptychography reconstructions were carried out using 500x500 pixels from the detector, resulting in a pixel size of 27.9 nm. The ptychoshelves package \cite{wakonig2020ptychoshelves} was used with 400 iterations of difference map \cite{thibault2008high} followed by maximum likelihood refinement~\citep{thibault2012maximum, odstrvcil2018iterative}. For both PXCT and PyXL projections were post-processed and aligned following the methods in~\citep{odstrvcil2019alignment}.

\textbf{Dataset composition and usage.} The PXCT data comprises 10 tomograms of mouse brain pillars, each containing rich ultrastructural detail including neurons, synapses, and fine neuropil textures. Voxel sizes range from 38 nm to 81 nm, with volumetric dimensions varying across samples, e.g., $512 \times 768 \times 768$ voxels, $400 \times 768 \times 768$ voxels, $256 \times 896 \times 896$ voxels, down to $128 \times 736 \times 736$ voxels. For the training, slices were randomly sampled from the first nine tomograms, separately for axial ($736 \times 736$ voxels), sagittal ($128 \times 736$ voxels), and coronal ($128 \times 736$ voxels) orientations to enable multi-view learning. For the inference, we generate laminography projections from the remaining tomogram of PXCT with $128 \times 736 \times 736$ voxels to do simulation, and use real laminography projections from PyXL for experimental validation.

During simulation, laminographic projections were synthetically generated from the tomography volume using a laminography forward operator, matching the acquisition geometry. In particular, projections were simulated at a fixed tilt angle of $\theta = 61^\circ$, reproducing the characteristic missing-cone sampling in Fourier space. For the given object thickness ($T \approx 4.7\,\mu\mathrm{m}$) and spatial resolution ($\Delta r \approx 37\,\mathrm{nm}$), the laminographic sampling criterion predicts that $N = 723$ projections are required to achieve Nyquist angular sampling, according to $ N = \pi \frac{T}{\Delta r} \tan \theta$ \cite{holler2019three}.
In contrast, we simulated only 360 uniformly spaced projections, corresponding to an angularly undersampled regime. This is evident in the gaps shown in the Fourier-domain gaps observed for FBP in Fig.~\ref{sim_compare}.

Despite this, LUCID yields stable and structurally faithful reconstructions, highlighting the ability of diffusion-based priors to compensate for incomplete angular sampling beyond the limits imposed by classical reconstruction theory.

\textbf{Evaluation.} 
To quantitatively assess reconstruction fidelity in the image domain, we use PSNR, SSIM, RMSE. Let $I$ denote the GT volume
and $\hat I$ the reconstructed volume, both defined on the voxel set
$\Omega$, where $|\Omega|$ denotes the total
number of voxels in the 3D volume. All metrics are computed over the 3D
brain volume.

We first define the mean squared error (MSE) as
\begin{equation}
    \mathrm{MSE}(I,\hat I)
    = \frac{1}{|\Omega|} \sum_{i \in \Omega} \bigl(I_i - \hat I_i\bigr)^2 .
\end{equation}
The dynamic range $L$ is estimated from the GT as
\begin{equation}
    L = \max_{i \in \Omega} I_i - \min_{i \in \Omega} I_i .
\end{equation}

The PSNR used in our experiments is then given by
\begin{equation}
    \mathrm{PSNR}(I,\hat I)
    = 20 \log_{10} \left( \frac{L}{\sqrt{\mathrm{MSE}(I,\hat I)}} \right) ,
\end{equation}
where higher values indicate better reconstruction quality.

The RMSE is defined as
\begin{equation}
    \mathrm{RMSE}(I,\hat I)
    = \sqrt{\mathrm{MSE}(I,\hat I)}
    = \sqrt{ \frac{1}{|\Omega|} \sum_{i \in \Omega} \bigl(I_i - \hat I_i\bigr)^2 } .
\end{equation}

For SSIM, we adopt the global formulation consistent with our
implementation~\citep{wang2003multiscale}. Let
\begin{equation}
    \mu_I   = \frac{1}{|\Omega|} \sum_{i \in \Omega} I_i, \qquad
    \mu_{\hat I} = \frac{1}{|\Omega|} \sum_{i \in \Omega} \hat I_i ,
\end{equation}
\begin{equation}
    \sigma_I^2   = \frac{1}{|\Omega|} \sum_{i \in \Omega} (I_i - \mu_I)^2, \qquad
    \sigma_{\hat I}^2 = \frac{1}{|\Omega|} \sum_{i \in \Omega} (\hat I_i - \mu_{\hat I})^2 ,
\end{equation}
\begin{equation}
    \sigma_{I\hat I}
    = \frac{1}{|\Omega|} \sum_{i \in \Omega}
      (I_i - \mu_I)(\hat I_i - \mu_{\hat I}) .
\end{equation}
The data range $L$ defined above is used to construct the SSIM stabilisation
constants
\begin{equation}
    C_1 = (k_1 L)^2, \qquad C_2 = (k_2 L)^2,
\end{equation}
with fixed parameters $k_1 = 0.01$ and $k_2 = 0.03$. The structural similarity
index between $I$ and $\hat I$ is thus
\begin{equation}
    \mathrm{SSIM}(I,\hat I)
    = \frac{(2 \mu_I \mu_{\hat I} + C_1)(2 \sigma_{I\hat I} + C_2)}
           {(\mu_I^2 + \mu_{\hat I}^2 + C_1)(\sigma_I^2 + \sigma_{\hat I}^2 + C_2)} .
\end{equation}
Higher SSIM values indicate greater perceptual similarity between the
reconstruction and the GT.

To evaluate spectral completeness, especially in the missing-cone region, we compute three Fourier-domain metrics.

\paragraph{In-cone spectral energy.}
\begin{equation}
E_{\mathrm{cone}}
=
\frac{\sum_{\mathbf{k}\in\mathcal{C}} |F_{\hat I}(\mathbf{k})|^2}
     {\sum_{\mathbf{k}\in\mathcal{C}} |F_I(\mathbf{k})|^2},
\end{equation}
where $\mathcal{C}$ denotes the missing-cone region in Fourier space.

\paragraph{In-cone energy difference.}
\begin{equation}
\Delta E_{\mathrm{cone}}
=
\frac{%
\sum_{\mathbf{k}\in\mathcal{C}}
\big(|F_{\hat I}(\mathbf{k})|^2 - |F_I(\mathbf{k})|^2\big)
}{%
\sum_{\mathbf{k}\in\mathcal{C}} |F_I(\mathbf{k})|^2
}.
\end{equation}

\paragraph{Cone spectral correlation.}
\begin{equation}
\mathrm{CSF}_{\mathrm{cone}}
=
\frac{\langle |F_{\hat I}|, |F_I|\rangle_{\mathcal{C}}}
     {\|F_{\hat I}\|_{\mathcal{C}} \cdot \|F_I\|_{\mathcal{C}}},
\end{equation}
which ranges from 0 to 1, with higher values indicating closer agreement with the GT.

Together, these spatial and spectral metrics provide a comprehensive evaluation of reconstruction fidelity and Fourier-space completion.

\textbf{Training details} As for model architecture, the diffusion model is based on U-Net with an encoder and decoder consisting of resnet blocks~\citep{ronneberger2015u}. We trained diffusion priors using AdamW with a cosine learning-rate schedule and 500 warm-up steps. The training data were extracted from axial, sagittal, and coronal slices of the datasets. In practice, two diffusion models were trained: one using axial slices and one using the combined sagittal and coronal slices. Because sagittal and coronal views share identical spatial dimensions and similar image statistics, a single model was trained on both orientations and subsequently applied to each view during inference.
Batch sizes were set to 10 for axial slices and 60 for sagittal/coronal slices to balance memory usage and training efficiency. Smaller batches used for the larger axial slices and larger batches used for the sagittal/coronal views.
We augment the data by random rotation and scaling during the training phase to reduce overfitting. Data augmentation is a process to synthetically generate additional training samples for the purpose of avoiding over-fitting and increasing robustness in the image domain. The implementation of all methods in this work is based on the PyTorch library, and all experiments are run on a single NVIDIA A100-PCIE.

\subsection{Inverse problem in X-ray laminography}

Reconstructing a volumetric object from X-ray projections is an inverse problem.
Given an unknown 3D object $x \in \mathbb{R}^n$, the measurement process is modelled as
\begin{equation}
    y = \mathcal{A}(x) + \eta,
\end{equation}
where $\mathcal{A}:\mathbb{R}^n \rightarrow \mathbb{R}^m$ denotes the forward projection operator which is given by the acquisition geometry, and $\eta$ represents measurement noise.  
In conventional CT, $\mathcal{A}$ corresponds to a Radon transform and is well-conditioned when sufficient angular coverage is available.

In laminography, the rotation axis has an angle $\beta < 90^\circ$ relative to the beam propagation direction, as opposed to the conventional $90^\circ$ of CT. This geometry is advantageous for imaging specimens that are extended in 2D. Despite this favourable measurement geometry laminography produces a characteristic deficit in the Fourier domain: in contrast to computed tomography, which achieves complete reciprocal-space coverage, laminography introduces a cone-shaped region of missing frequencies. The resulting incomplete sampling makes the laminographic operator $\mathcal{A}$  ill-conditioned, leading in the reconstructions to axial elongation, mixing between adjacent layers, and highly directional blurring artifacts.
In this work, we address this challenge by integrating a learned generative prior with explicit physical constraints, forming a unified, data-consistent diffusion framework for 3D brain imaging.

\subsection{Denoising diffusion probabilistic models}

Denoising diffusion probabilistic models (DDPMs)~\citep{ho2020denoising} learn complex data distributions via a gradual forward noising process and a learned reverse denoising process.
Let $\mathbf{x}_0 \in \mathbb{R}^d$ denote a clean data sample, and let $\{\mathbf{x}_t\}_{t=1}^T$ denote a sequence of latent variables indexed by the diffusion time step $t \in \{1,\dots,T\}$.
The forward process is defined as a Markov chain $q$, where each transition adds Gaussian noise:
\begin{equation}
q(\mathbf{x}_t | \mathbf{x}_{t-1})
= \mathcal{N}\!\left(\mathbf{x}_t; \sqrt{1-\beta_t}\,\mathbf{x}_{t-1},\, \beta_t \mathbf{I}\right),
\end{equation}
where $\mathcal{N}(\cdot;\boldsymbol{\mu},\boldsymbol{\Sigma})$ denotes a multivariate normal distribution with mean $\boldsymbol{\mu}$ and covariance $\boldsymbol{\Sigma}$,
$\mathbf{I}\in\mathbb{R}^{d\times d}$ is the identity matrix, and $\{\beta_t\}_{t=1}^T$ is a predefined variance schedule controlling the noise magnitude at each step.

By recursively composing the linear Gaussian transitions of the Markov chain, the marginal distribution of $\mathbf{x}_t$ conditioned on the original sample $\mathbf{x}_0$ admits a closed-form expression:
\begin{equation}
q(\mathbf{x}_t | \mathbf{x}_0)
= \mathcal{N}\!\left(\mathbf{x}_t; \sqrt{\alpha_t}\, \mathbf{x}_0,\,(1-\alpha_t)\mathbf{I}\right),
\quad \alpha_t = \prod_{j=1}^t (1-\beta_j).
\end{equation}

The reverse process $p_\theta$ is parameterized by a neural network $s_\theta$ that predicts the noise component at step $t$:
\begin{equation}
p_\theta(\mathbf{x}_{t-1} | \mathbf{x}_t)
= \mathcal{N}\!\left(
\mathbf{x}_{t-1};\,
\frac{1}{\sqrt{1-\beta_t}}\big( \mathbf{x}_t + \beta_t s_\theta(\mathbf{x}_t,t)\big),\,
\beta_t\mathbf{I}
\right).
\end{equation}

During inference, a sample is generated by initializing $\mathbf{x}_T \sim \mathcal{N}(0, \mathbf{I})$ and iteratively applying the reverse transitions.
In LUCID, diffusion models serve as learned priors of clean brain structures, providing biologically and statistically coherent regularization for the ill-posed laminography reconstruction.

\subsection{LUCID: Laminography with Unified Consistent Diffusion}

To incorporate 3D anatomical coherence while maintaining computational tractability, LUCID adopts a multi-view diffusion strategy.  
Diffusion priors are learned from axial, sagittal, and coronal slices. In practice, two DDPMs are trained: one on axial slices and one on the combined sagittal and coronal slices, which share identical dimensions.
Each model learns the slice distribution in its respective orientation, capturing complementary structural cues across views.

\subsubsection{Multi-view diffusion prior}
At diffusion step $t$, LUCID performs stochastic denoising updates using view-specific diffusion models. 
For each orientation (axial, sagittal, or coronal), the noisy sample is updated as
\begin{equation}
\mathbf{x}_{t-1}^{\mathrm{view}}
=
\frac{1}{\sqrt{1-\beta_t}}
\Big(
\mathbf{x}_t^{\mathrm{view}}
+
\beta_t\, s_{\theta,\mathrm{view}}(\mathbf{x}_t^{\mathrm{view}}, t)
\Big)
+
\sqrt{\beta_t}\,\mathbf{z}_t,
\end{equation}
where $s_{\theta,\mathrm{view}}$ denotes the view-specific denoiser that predicts the noise component,
$\beta_t$ is the diffusion variance schedule, and $\mathbf{z}_t \sim \mathcal{N}(0,\mathbf{I})$.
The denoisers are applied in one view using the corresponding view-specific diffusion model, then refined by the laminography data consistency module. The resulting volume is passed to the next view-specific denoising step. The views are cycled every three steps, so the axial, sagittal, and coronal priors are incorporated sequentially.

\vspace{0.5em}
\noindent
To enable the application of physical measurement operators, we further compute a noise-free estimate of the clean volume using Tweedie’s formula. Specifically, from the current noisy state $\mathbf{x}_t$, we form
\begin{equation}
\hat{\mathbf{x}}_{0}
=
\frac{1}{\sqrt{\bar{\alpha}_t}}
\Big(
\mathbf{x}_t
-
\sqrt{1-\bar{\alpha}_t}\,\boldsymbol{\epsilon}_{\theta}(\mathbf{x}_t, t)
\Big),
\end{equation}
where $\bar{\alpha}_t = \prod_{j=1}^{t}(1-\beta_j)$ and $\boldsymbol{\epsilon}_{\theta}$ denotes the aggregated noise prediction obtained from the multi-view denoisers.
This estimate serves as a deterministic approximation of the underlying clean volume at step $t$.

\subsubsection{Laminography projection-domain data consistency}
To enforce consistency with the measured laminography data, projection-domain constraints are imposed on the noise-free estimate $\hat{\mathbf{x}}_{0}$.
Simulated laminography projections are obtained via the forward operator
\begin{equation}
\hat{\mathbf{y}} = \mathcal{A}_{\theta}(\hat{\mathbf{x}}_{0}),
\end{equation}
where $\mathcal{A}_{\theta}$ denotes the laminography projection operator at tilt angle $\theta$.
The simulated projections $\hat{\mathbf{y}}$ are compared with the experimentally measured projections $\mathbf{y}_{\mathrm{measured}}$,
and the estimate is refined by minimizing the data-fidelity term through gradient descent:
\begin{equation}
\hat{\mathbf{x}}_{0}^{(k+1)}
=
\hat{\mathbf{x}}_{0}^{(k)}
-
\eta
\nabla_{\hat{\mathbf{x}}_{0}^{(k)}}
\big\|
\mathcal{A}_{\theta}(\hat{\mathbf{x}}_{0}^{(k)}) - \mathbf{y}_{\mathrm{measured}}
\big\|_2^2,
\end{equation}
where $\eta$ is the step size and $k$ indexes the inner data-consistency iterations.
This projection-domain correction enforces adherence to the physical measurements while preserving the anatomical priors provided by the diffusion model.

\subsubsection{Domain translation for real laminography inference}

To bridge the gap between simulated tomography-domain training data and real laminography measurements, we employ a reversible domain translation module. A forward mapping $f_{\mathrm{trans}}$ aligns real reconstructions to the tomography-like intensity manifold via core-region histogram matching and global monotonic quantile interpolation. After diffusion-based denoising, an approximate inverse mapping $f_{\mathrm{inv}}$ restores the output to the laminography domain by applying the inverse quantile transform, removing auxiliary padding, and reintegrating data-consistent regions. This bidirectional mapping enables stable diffusion inference while preserving compatibility with the physical laminography operator.

\subsubsection{Iterative sampling loop}

The full reconstruction process alternates between (i) multi-view diffusion prior module, (ii) laminography projection-domain consistency module, and (iii) controlled noise reintroduction.
Iterating these steps from $t=T$ to $t=1$ yields a final estimate $\mathbf{x}_0$ that is simultaneously physically consistent and anatomically plausible.


\section{Data availability}

All data supporting the findings described in this manuscript are available in the article and in the Supplementary Information.
A supplementary structured table including metadata supporting the measurements presented is available via Zenodo. The PXCT data from pillar-shaped samples are available at \href{https://doi.org/10.5281/zenodo.16362800}{Zenodo}. And the PyXL brain datasets are available for preview at \href{https://zenodo.org/records/20268049?preview=1&token=eyJhbGciOiJIUzUxMiIsImlhdCI6MTc3OTI4NDM0NCwiZXhwIjoxNzk4Njc1MTk5fQ.eyJpZCI6IjI2NWNjYmM1LWFlOGItNGMyNS1iZDBjLWEyZDUyYTgwMWJiYSIsImRhdGEiOnt9LCJyYW5kb20iOiJhYjc5ZGQ0NTdhODRjYzAyYzUyYmQ5ZDc2NDNkMjU2MCJ9.1NmpxzwP4xQEI97eYJvfocbT6gr93g9X1_Y5hFc8saEvw1CT-rtYuj7OCp3_Ap4q8qWGZxWpTT3PvKppYieU_g}{Zenodo}.
All datasets can be viewed and downloaded from the links provided in Supplementary Information 1.

\section{Code availability}

The code of LUCID is available under MIT license at Github \url{https://github.com/Athenaxr/LUCID.git}.The exact version used in this work
(v1.0.0) has been archived at Zenodo:
\url{https://doi.org/10.5281/zenodo.20430172}. The ptychography reconstruction code is available from \url{https://www.psi.ch/en/sls/csaxs/software} (license:\url{https://www.psi.ch/sites/default/files/import/sls/csaxs/ComputingEN/License.txt)}. 

\bibliography{sn-bibliography}

@article{pallotto2015extracellular,
  title={Extracellular space preservation aids the connectomic analysis of neural circuits},
  author={Pallotto, Marta and Watkins, Paul V and Fubara, Boma and Singer, Joshua H and Briggman, Kevin L},
  journal={Elife},
  volume={4},
  pages={e08206},
  year={2015},
  publisher={eLife Sciences Publications, Ltd}
}

@article{thibault2012maximum,
  title={Maximum-likelihood refinement for coherent diffractive imaging},
  author={Thibault, Pierre and Guizar-Sicairos, Manuel},
  journal={New Journal of Physics},
  volume={14},
  number={6},
  pages={063004},
  year={2012},
  publisher={IOP Publishing}
}

@article{walton1979lead,
  title={Lead asparate, an en bloc contrast stain particularly useful for ultrastructural enzymology.},
  author={Walton, Judie},
  journal={Journal of Histochemistry \& Cytochemistry},
  volume={27},
  number={10},
  pages={1337--1342},
  year={1979},
  publisher={SAGE Publications Sage CA: Los Angeles, CA}
}

@article{helmstaedter2026synaptic,
  title={Synaptic-resolution connectomics: towards large brains and connectomic screening},
  author={Helmstaedter, Moritz},
  journal={Nature Reviews Neuroscience},
  volume={27},
  number={2},
  pages={101--120},
  year={2026},
  publisher={Nature Publishing Group UK London}
}

@article{holler2018omny,
  title={OMNY—a tOMography nano crYo stage},
  author={Holler, Mirko and Raabe, Joerg and Diaz, Ana and Guizar-Sicairos, Manuel and Wepf, R and Odstrcil, M and Shaik, Farooque R and Panneels, Val{\'e}rie and Menzel, Andreas and Sarafimov, Blagoj and others},
  journal={Review of Scientific Instruments},
  volume={89},
  number={4},
  year={2018},
  publisher={AIP Publishing}
}

@article{wakonig2020ptychoshelves,
  title={PtychoShelves, a versatile high-level framework for high-performance analysis of ptychographic data},
  author={Wakonig, Klaus and Stadler, H-C and Odstr{\v{c}}il, Michal and Tsai, Esther HR and Diaz, Ana and Holler, Mirko and Usov, Ivan and Raabe, J{\"o}rg and Menzel, Andreas and Guizar-Sicairos, Manuel},
  journal={Applied Crystallography},
  volume={53},
  number={2},
  pages={574--586},
  year={2020},
  publisher={International Union of Crystallography}
}

@inproceedings{lee2023improving,
  title={Improving 3D imaging with pre-trained perpendicular 2D diffusion models},
  author={Lee, Suhyeon and Chung, Hyungjin and Park, Minyoung and Park, Jonghyuk and Ryu, Wi-Sun and Ye, Jong Chul},
  booktitle={Proceedings of the IEEE/CVF international conference on computer vision},
  pages={10710--10720},
  year={2023}
}

@article{guizar2014high,
  title={High-throughput ptychography using Eiger: scanning X-ray nano-imaging of extended regions},
  author={Guizar-Sicairos, Manuel and Johnson, Ian and Diaz, Ana and Holler, Mirko and Karvinen, Petri and Stadler, Hans-Christian and Dinapoli, Roberto and Bunk, Oliver and Menzel, Andreas},
  journal={Optics express},
  volume={22},
  number={12},
  pages={14859--14870},
  year={2014},
  publisher={Optical Society of America}
}

@article{huang2014optimization,
  title={Optimization of overlap uniformness for ptychography},
  author={Huang, Xiaojing and Yan, Hanfei and Harder, Ross and Hwu, Yeukuang and Robinson, Ian K and Chu, Yong S},
  journal={Optics Express},
  volume={22},
  number={10},
  pages={12634--12644},
  year={2014},
  publisher={Optical Society of America}
}

@article{odstrvcil2019towards,
  title={Towards optimized illumination for high-resolution ptychography},
  author={Odstr{\v{c}}il, Michal and Lebugle, Maxime and Guizar-Sicairos, Manuel and David, Christian and Holler, Mirko},
  journal={Optics express},
  volume={27},
  number={10},
  pages={14981--14997},
  year={2019},
  publisher={Optical Society of America}
}

@article{bosch2025nondestructive,
  title={Nondestructive X-ray tomography of brain tissue ultrastructure},
  author={Bosch, Carles and Aidukas, Tomas and Holler, Mirko and Pacureanu, Alexandra and M{\"u}ller, Elisabeth and Peddie, Christopher J and Zhang, Yuxin and Cook, Phil and Collinson, Lucy and Bunk, Oliver and others},
  journal={Nature Methods},
  pages={1--8},
  year={2025},
  publisher={Nature Publishing Group US New York}
}

@article{yin2024survey,
  title={A survey on multimodal large language models},
  author={Yin, Shukang and Fu, Chaoyou and Zhao, Sirui and Li, Ke and Sun, Xing and Xu, Tong and Chen, Enhong},
  journal={National Science Review},
  volume={11},
  number={12},
  pages={nwae403},
  year={2024},
  publisher={Oxford University Press}
}

@article{song2019generative,
  title={Generative modeling by estimating gradients of the data distribution},
  author={Song, Yang and Ermon, Stefano},
  journal={Advances in neural information processing systems},
  volume={32},
  year={2019}
}

@article{arslan2006reducing,
  title={Reducing the missing wedge: High-resolution dual axis tomography of inorganic materials},
  author={Arslan, Ilke and Tong, Jenna R and Midgley, Paul A},
  journal={Ultramicroscopy},
  volume={106},
  number={11-12},
  pages={994--1000},
  year={2006},
  publisher={Elsevier}
}

@article{bartesaghi2008classification,
  title={Classification and 3D averaging with missing wedge correction in biological electron tomography},
  author={Bartesaghi, Alberto and Sprechmann, Pablo and Liu, Jun and Randall, Gregory and Sapiro, Guillermo and Subramaniam, Sriram},
  journal={Journal of structural biology},
  volume={162},
  number={3},
  pages={436--450},
  year={2008},
  publisher={Elsevier}
}

@article{odstrvcil2018iterative,
  title={Iterative least-squares solver for generalized maximum-likelihood ptychography},
  author={Odstr{\v{c}}il, Michal and Menzel, Andreas and Guizar-Sicairos, Manuel},
  journal={Optics express},
  volume={26},
  number={3},
  pages={3108--3123},
  year={2018},
  publisher={Optical Society of America}
}

@article{odstrcil2019ab,
  title={Ab initio nonrigid X-ray nanotomography},
  author={Odstrcil, Michal and Holler, Mirko and Raabe, J{\"o}rg and Sepe, Alessandro and Sheng, Xiaoyuan and Vignolini, Silvia and Schroer, Christian G and Guizar-Sicairos, Manuel},
  journal={Nature communications},
  volume={10},
  number={1},
  pages={2600},
  year={2019},
  publisher={Nature Publishing Group UK London}
}

@article{odstrvcil2019alignment,
  title={Alignment methods for nanotomography with deep subpixel accuracy},
  author={Odstr{\v{c}}il, Michal and Holler, Mirko and Raabe, J{\"o}rg and Guizar-Sicairos, Manuel},
  journal={Optics Express},
  volume={27},
  number={25},
  pages={36637--36652},
  year={2019},
  publisher={Optical Society of America}
}

@article{knott2008serial,
  title={Serial section scanning electron microscopy of adult brain tissue using focused ion beam milling},
  author={Knott, Graham and Marchman, Herschel and Wall, David and Lich, Ben},
  journal={Journal of Neuroscience},
  volume={28},
  number={12},
  pages={2959--2964},
  year={2008},
  publisher={Soc Neuroscience}
}

@article{merchan2009counting,
  title={Counting synapses using FIB/SEM microscopy: a true revolution for ultrastructural volume reconstruction},
  author={Merchan-Perez, Angel and Rodriguez, Jos{\'e}-Rodrigo and AlonsoNanclares, Lidia and Schertel, Andreas and DeFelipe, Javier},
  journal={Frontiers in Neuroanatomy},
  volume={3},
  pages={944},
  year={2009},
  publisher={Frontiers}
}

@article{holler2019three,
  title={Three-dimensional imaging of integrated circuits with macro-to nanoscale zoom},
  author={Holler, Mirko and Odstrcil, Michal and Guizar-Sicairos, Manuel and Lebugle, Maxime and M{\"u}ller, Elisabeth and Finizio, Simone and Tinti, Gemma and David, Christian and Zusman, Joshua and Unglaub, Walter and others},
  journal={Nature Electronics},
  volume={2},
  number={10},
  pages={464--470},
  year={2019},
  publisher={Nature Publishing Group UK London}
}

@article{holler2020lamni,
  title={LamNI--an instrument for X-ray scanning microscopy in laminography geometry},
  author={Holler, Mirko and Odstr{\v{c}}il, Michal and Guizar-Sicairos, Manuel and Lebugle, Maxime and Frommherz, Ulrich and Lachat, Thierry and Bunk, Oliver and Raabe, Joerg and Aeppli, Gabriel},
  journal={Journal of synchrotron radiation},
  volume={27},
  number={3},
  pages={730--736},
  year={2020},
  publisher={International Union of Crystallography}
}

@article{dierolf2010ptychographic,
  title={Ptychographic X-ray computed tomography at the nanoscale},
  author={Dierolf, Martin and Menzel, Andreas and Thibault, Pierre and Schneider, Philipp and Kewish, Cameron M and Wepf, Roger and Bunk, Oliver and Pfeiffer, Franz},
  journal={Nature},
  volume={467},
  number={7314},
  pages={436--439},
  year={2010},
  publisher={Nature Publishing Group UK London}
}

@article{thibault2008high,
  title={High-resolution scanning x-ray diffraction microscopy},
  author={Thibault, Pierre and Dierolf, Martin and Menzel, Andreas and Bunk, Oliver and David, Christian and Pfeiffer, Franz},
  journal={Science},
  volume={321},
  number={5887},
  pages={379--382},
  year={2008},
  publisher={American Association for the Advancement of Science}
}

@article{kang2023accelerated,
  title={Accelerated deep self-supervised ptycho-laminography for three-dimensional nanoscale imaging of integrated circuits},
  author={Kang, Iksung and Jiang, Yi and Holler, Mirko and Guizar-Sicairos, Manuel and Levi, Anthony FJ and Klug, Jeffrey and Vogt, Stefan and Barbastathis, George},
  journal={Optica},
  volume={10},
  number={8},
  pages={1000--1008},
  year={2023},
  publisher={Optica Publishing Group}
}

@article{ho2020denoising,
  title={Denoising diffusion probabilistic models},
  author={Ho, Jonathan and Jain, Ajay and Abbeel, Pieter},
  journal={Advances in neural information processing systems},
  volume={33},
  pages={6840--6851},
  year={2020}
}

@article{song2020score,
  title={Score-based generative modeling through stochastic differential equations},
  author={Song, Yang and Sohl-Dickstein, Jascha and Kingma, Diederik P and Kumar, Abhishek and Ermon, Stefano and Poole, Ben},
  journal={arXiv preprint arXiv:2011.13456},
  year={2020}
}

@article{abbott2020mind,
  title={The mind of a mouse},
  author={Abbott, Larry F and Bock, Davi D and Callaway, Edward M and Denk, Winfried and Dulac, Catherine and Fairhall, Adrienne L and Fiete, Ila and Harris, Kristen M and Helmstaedter, Moritz and Jain, Viren and others},
  journal={Cell},
  volume={182},
  number={6},
  pages={1372--1376},
  year={2020},
  publisher={Elsevier}
}

@inproceedings{ronneberger2015u,
  title={U-net: Convolutional networks for biomedical image segmentation},
  author={Ronneberger, Olaf and Fischer, Philipp and Brox, Thomas},
  booktitle={International Conference on Medical image computing and computer-assisted intervention},
  pages={234--241},
  year={2015},
  organization={Springer}
}

@inproceedings{wang2003multiscale,
  title={Multiscale structural similarity for image quality assessment},
  author={Wang, Zhou and Simoncelli, Eero P and Bovik, Alan C},
  booktitle={The thrity-seventh asilomar conference on signals, systems \& computers, 2003},
  volume={2},
  pages={1398--1402},
  year={2003},
  organization={Ieee}
}

@article{helfen2009phase,
  title={Phase-contrast and holographic computed laminography},
  author={Helfen, L and Baumbach, T and Cloetens, P and Baruchel, J},
  journal={Applied Physics Letters},
  volume={94},
  number={10},
  pages={104103},
  year={2009},
  publisher={American Institute of Physics}
}

@article{zou2025artifact,
  title={Artifact reduction in rotational computed laminography using a deep learning method},
  author={Zou, Xiang and Shi, Wuliang and Du, Muge and Xing, Yuxiang},
  journal={Optics and Lasers in Engineering},
  volume={187},
  pages={108881},
  year={2025},
  publisher={Elsevier}
}

@article{que2012computed,
  title={Computed laminography and reconstruction algorithm},
  author={Que, Jie-Min and Cao, Da-Quan and Zhao, Wei and Tang, Xiao and Sun, Cui-Li and Wang, Yan-Fang and Wei, Cun-Feng and Shi, Rong-Jian and Wei, Long and Yu, Zhong-Qiang and others},
  journal={Chinese Physics C},
  volume={36},
  number={8},
  pages={777},
  year={2012},
  publisher={IOP Publishing}
}

@article{lu2023anisotropic,
  title={An anisotropic alternating regularization-based reconstruction algorithm for cone beam computed laminography},
  author={Lu, Jing and Liu, Yi and Zhang, Pengcheng and Li, Zhiyuan and Yang, Min and Gui, Zhiguo},
  journal={NDT \& E International},
  volume={138},
  pages={102898},
  year={2023},
  publisher={Elsevier}
}

@article{jia2023multi,
  title={The multi-scale fusion reconstruction algorithm of CT and CL},
  author={Jia, Tong and Wei, Cunfeng and Zhu, Min and Shi, Rongjian and Wang, Zhe and Cui, Xindong and Liu, Baodong},
  journal={Physica Scripta},
  volume={98},
  number={10},
  pages={105114},
  year={2023},
  publisher={IOP Publishing}
}

@article{ghandourah2023evaluation,
  title={Evaluation of welding imperfections with X-ray computed laminography for NDT inspection of carbon steel plates},
  author={Ghandourah, Emad E and Hamidi, Shahfuan Hanif A and Mohd Salleh, Khairul Anuar and Wahab, Mahamad Noor and Banoqitah, Essam Mohammed and Alhawsawi, Abdulsalam Mohammed and Moustafa, Essam B},
  journal={Journal of Nondestructive Evaluation},
  volume={42},
  number={3},
  pages={77},
  year={2023},
  publisher={Springer}
}

@article{xu2012comparison,
  title={Comparison of image quality in computed laminography and tomography},
  author={Xu, Feng and Helfen, Lukas and Baumbach, Tilo and Suhonen, Heikki},
  journal={Optics Express},
  volume={20},
  number={2},
  pages={794--806},
  year={2012},
  publisher={Optical Society of America}
}

@article{liu2026laminodiff,
  title={LaminoDiff: Artifact-Free Computed Laminography in Non-Destructive Testing via Diffusion Model},
  author={Liu, Tan and Shi, Liu and Peng, Binghuang and Jia, Tong and Xu, Xiaoling and Liu, Baodong and Liu, Qiegen},
  journal={arXiv preprint arXiv:2601.07254},
  year={2026}
}

@article{du2025x,
  title={X-ray computed laminography: A brief review of mechanisms, reconstruction, applications and perspectives},
  author={Du, Wenjia and Iacoviello, Francesco and Mirza, Mateen and Zhou, Shangwei and Bu, Junfu and Feng, Shikang and Grant, Patrick S and Jervis, Rhodri and Brett, Dan JL and Shearing, Paul R},
  journal={Materials Today},
  year={2025},
  publisher={Elsevier}
}

@article{ding2019joint,
  title={A joint deep learning model to recover information and reduce artifacts in missing-wedge sinograms for electron tomography and beyond},
  author={Ding, Guanglei and Liu, Yitong and Zhang, Rui and Xin, Huolin L},
  journal={Scientific reports},
  volume={9},
  number={1},
  pages={12803},
  year={2019},
  publisher={Nature Publishing Group UK London}
}

@article{wiedemann2024deep,
  title={A deep learning method for simultaneous denoising and missing wedge reconstruction in cryogenic electron tomography},
  author={Wiedemann, Simon and Heckel, Reinhard},
  journal={Nature Communications},
  volume={15},
  number={1},
  pages={8255},
  year={2024},
  publisher={Nature Publishing Group UK London}
}

@article{liu2022isotropic,
  title={Isotropic reconstruction for electron tomography with deep learning},
  author={Liu, Yun-Tao and Zhang, Heng and Wang, Hui and Tao, Chang-Lu and Bi, Guo-Qiang and Zhou, Z Hong},
  journal={Nature communications},
  volume={13},
  number={1},
  pages={6482},
  year={2022},
  publisher={Nature Publishing Group UK London}
}

@article{beck2025situ,
  title={In-Situ Ptychographic Nanotomography Captures Activation, Mobility, and Deactivation of Supported Catalysts},
  author={Beck, Arik and Holler, Mirko and Aidukas, Tomas and Menzel, Andreas and Guizar-Sicairos, Manuel and van Bokhoven, Jeroen A and Ihli, Johannes},
  journal={preprint},
  year={2025}
}

@article{yu2018three,
  title={Three-dimensional localization of nanoscale battery reactions using soft X-ray tomography},
  author={Yu, Young-Sang and Farmand, Maryam and Kim, Chunjoong and Liu, Yijin and Grey, Clare P and Strobridge, Fiona C and Tyliszczak, Tolek and Celestre, Rich and Denes, Peter and Joseph, John and others},
  journal={Nature communications},
  volume={9},
  number={1},
  pages={921},
  year={2018},
  publisher={Nature Publishing Group UK London}
}

@article{aidukas2024high,
  title={High-performance 4-nm-resolution X-ray tomography using burst ptychography},
  author={Aidukas, Tomas and Phillips, Nicholas W and Diaz, Ana and Poghosyan, Emiliya and M{\"u}ller, Elisabeth and Levi, Anthony FJ and Aeppli, Gabriel and Guizar-Sicairos, Manuel and Holler, Mirko},
  journal={Nature},
  volume={632},
  number={8023},
  pages={81--88},
  year={2024},
  publisher={Nature Publishing Group UK London}
}

@article{nikitin2024laminography,
  title={Laminography as a tool for imaging large-size samples with high resolution},
  author={Nikitin, Viktor and Wildenberg, Gregg and Mittone, Alberto and Shevchenko, Pavel and Deriy, Alex and De Carlo, Francesco},
  journal={Synchrotron Radiation},
  volume={31},
  number={4},
  pages={851--866},
  year={2024},
  publisher={International Union of Crystallography}
}

@article{laugros2025self,
  title={Self-supervised image restoration in coherent x-ray neuronal microscopy},
  author={Laugros, Alfred and Cloetens, Peter and Bosch, Carles and Schoonhoven, Richard and Pavlovic, Liam and Kuan, Aaron T and Livingstone, Jayde and Zhang, Yuxin and Kim, Minsu and Hendriksen, Allard and others},
  journal={BioRxiv},
  pages={2025--02},
  year={2025},
  publisher={Cold Spring Harbor Laboratory}
}








\section{Acknowledgements}

We acknowledge the Paul Scherrer Institute, Villigen PSI, Switzerland, for the provision of synchrotron radiation beamtime at the cSAXS beamline of the Swiss Light Source. The work of L.B was supported by the Swiss Data Science Center (SDSC) under the CHIP project grant no. C22-11L. A.L.L. and N.W.P were supported by the European Union’s Horizon 2020 research and innovation program under the Marie Sklodowska-Curie grant agreement no.884104 (PSI-FELLOW-III-3i). T.A. was supported by funding from the Swiss National Science Foundation (SNF), project number 200021\_196898. This work was also supported by the Francis Crick Institute, which receives its core funding from Cancer Research UK (CC2036 to A.T.S.), the UK Medical Research Council (CC2036 to A.T.S.), and the Wellcome Trust (CC2036 and 110174/Z/15/Z to A.T.S.). It was also supported by a Physics of Life grant (EP/W024292/1) to A.T.S. and A.P. funded by EPSRC and Wellcome.  

\section{Author contributions}
W.F., L.B., and M.G.-S. conceived the research project. W.F. and L.B. developed the core algorithms. A.L.L. provided technical support throughout the project. C.B.P. contributed biological expertise and guided the validation of neuroanatomical structures. A.D., A.W., A.T.S., A.P., M.H., N.W.P., T.A, Y.Z. and M.G.-S. secured beamtime and performed the experiments, acquiring the brain imaging datasets and performing ptychographic and tomographic reconstructions.

All authors contributed to the interpretation of results and provided critical feedback on the manuscript.

\end{document}